\documentclass[conference]{IEEEtran}
\IEEEoverridecommandlockouts
\usepackage[noadjust]{cite}

\usepackage{nicefrac}
\usepackage{booktabs}
\usepackage{amsmath,amssymb,amsfonts}
\usepackage{graphicx}
\usepackage[labelformat=simple]{subcaption}

\def\BibTeX{{\rm B\kern-.05em{\sc i\kern-.025em b}\kern-.08em
    T\kern-.1667em\lower.7ex\hbox{E}\kern-.125emX}}

\newcommand{\citet}[1]{\cite{#1}}
\newcommand{\citep}[1]{\cite{#1}}
    
\begin{document}

\title{Assessing the Reliability of Visual Explanations of\\Deep Models with Adversarial Perturbations\\
\thanks{We thank the partial support given by the Project: Models, Algorithms and Systems for the Web (grant FAPEMIG / PRONEX / MASWeb APQ-01400-14), and authors' individual grants and scholarships from CNPq, Fapemig and Kunumi. Tiago Pimentel is now at University of Cambridge.}
}

\author{\IEEEauthorblockN{Dan Valle}
\IEEEauthorblockA{
\textit{CS Dept@UFMG \& Wildlife Studios} \\
Belo Horizonte, Brazil\\
dan.valle@wildlifestudios.com}
\and
\IEEEauthorblockN{Tiago Pimentel}
\IEEEauthorblockA{
\textit{CS Dept@UFMG \& Kunumi} \\
Belo Horizonte, Brazil \\
tpimentelms@gmail.com}
\and
\IEEEauthorblockN{Adriano Veloso}
\IEEEauthorblockA{
\textit{CS Dept@UFMG} \\
Belo Horizonte, Brazil \\
adrianov@dcc.ufmg.br}
}

\maketitle

\begin{abstract}
The interest in complex deep neural networks for computer vision applications is increasing. This leads to the need for improving the interpretable capabilities of these models. Recent explanation methods present visualizations of the relevance of pixels from input images, thus enabling the direct interpretation of properties of the input that lead to a specific output. These methods produce maps of pixel importance, which are commonly evaluated by visual inspection. This means that the effectiveness of an explanation method is assessed based on human expectation instead of actual feature importance. Thus, in this work we propose an objective measure to evaluate the reliability of explanations of deep models. Specifically, our approach is based on changes in the network's outcome resulting from the perturbation of input images in an adversarial way. We present a comparison between widely-known explanation methods using our proposed approach. Finally, we also propose a straightforward application of our approach to clean relevance maps, creating more interpretable maps without any loss in essential explanation (as per our proposed measure).
\end{abstract}

\begin{IEEEkeywords}
Deep Networks, Explainability, Interpretability
\end{IEEEkeywords}

\section{Introduction}

The development of deep neural networks for computer vision applications is at the crossroad of two major trends. The first trend is associated with increasingly complex models leading to state-of-the-art performance in applications that vary from  healthcare \cite{miotto2017deep,shen2017deep} to economics  \cite{hartford2017deep}, to ranking~\cite{ref9}. The second trend is the increasingly perceived importance of transparency and accountability in a range of applications.

Deep neural networks do not provide insights into their complex behavior, and thus several methods attempt to unveil the factors contributing most to these networks black-box decisions \cite{ref12,ribeiro2016should,baehrens2010explain,fong2017interpretable}. These explanations are important to identify potential bias/problems in the training data~\cite{ref1}, to ensure compliance to existing regulations, and guarantee the model performs as expected \cite{cadamuro2016debugging}. This leads to an increase in interpretability~\cite{ref15}, making models more transparent and their predictions more understandable.

The explanation of deep models for computer vision applications is usually given in terms of interpretable visualizations of the relevance of pixels from input images (i.e., pixel relevance maps). Currently, the reliability of these explanations is largely assessed by visual inspection, and thus it is likely to be the case that existing explanation methods are being evaluated based purely on human expectation rather than on actual feature importance \cite{adebayo2018sanity}. As deep neural networks have special behaviors~\cite{ref14} and can be easily confused \cite{nguyen2015deep,ref6,goodfellow6572explaining}, intuitive visualizations can be misleading and different from the real importance given to features in a trained network.

An objective measure from which the quality of model explanations can be systematically assessed is largely lacking. In this paper we propose an approach for comparing the reliability of explanations. Our approach calculates a reliability measure based on changes in the model outcome resulting from adversarial perturbations of input images. Our specific contributions are summarized as follows:

\begin{itemize}
\item We introduce Adversarial Perturbation Explanation Measure (APEM), which evaluates pixel relevance maps by assuming model decisions must depend on its explanation \cite{bach-plos15, samek2017evaluating}. Therefore, the most relevant features in the input image are the ones that influence the most the output of the model. To calculate APEM, we probe the model outcome by perturbing the input image based on a relevance map, multiplied by the sign of the gradient, trying to maximize the error with minimal input modification.
\item We compare a variety of \emph{state-of-the-art} methods used for assessing feature importance in a controlled and standardized setting. We show how common practices in these explanation methods abdicate important information impacting models decisions in order to make visualizations more comprehensible and visually meaningful.
\item We show that it is also important to consider the low scores of relevance in order to avoid privileging explanation methods whose maps are restricted to a few concentrated high values. Thus, we use \emph{irrelevance} maps together with relevance ones, and we show that this balances the `precision' and `recall' of the resulting maps.
\item We present a new approach based on the proposed measure to clean relevance maps, making them more intuitive and understandable while keeping explanations reliable. This results in images with minimum noise while ensuring that no important explanation is lost.
\end{itemize}

\section{Background and Related Work}
\label{sec:relwork}

In this section we give an overview of methods focused on explaining deep network outcomes. Then, we present studies reporting how visualizations can be unreliable in some cases. Finally, we also discuss approaches for evaluating the quality of relevance maps generated by existing explanation methods.

\subsection{Explaining Deep Neural Network Decisions}

The large number of layers employed by deep models combined with their non-linearities makes it difficult to identify what is being considered in each decision. Some works attempt to understand individual neurons in deep neural networks by creating visualizations from higher level features in
Autoencoders~\cite{erhan2009visualizing, le2013building} and Deep Belief Networks~\cite{erhan2009visualizing}. These studies show how models can find patterns that are similar to what humans consider relevant in the domains analyzed.

Understanding what is important to the performance of deep models is essential to find situations in which they fail and to improve them. Authors in \cite{zeiler2014visualizing} created a visualization for each input image, which shows the patterns in the input that resulted in a specific activation on further layers. Other techniques tackle this problem by inverting the input images representations and analyzing their remaining information \cite{mahendran2015understanding, dosovitskiy2016inverting}.

Authors in \cite{simonyan2013deep} presented a gradient-based approach to interpret decisions of Convolutional Neural Networks (CNNs). They proposed saliency maps that are computed using the gradient of each class' score. This method generates visually noisy relevance maps, though. Smooth Grad \cite{smilkov2017smoothgrad} builds on their work and tackles the issue with noisy images. In order to do so, the authors created $n$ noisy copies from each input image and average the relevance maps calculated for them. Other works also use the gradient in different ways to achieve contrasting visualization results  \cite{springenberg2014striving, selvaraju2017grad,sundararajan2017axiomatic}. LIME \cite{ribeiro2016should}, on the other hand, explains predictions by learning an interpretable model locally around the input image.

A distinct group of model explanation techniques calculates relevance scores for pixels by applying a ``backpropagation'' method. It computes the relevance from the output back to the input image, assuring a layer-wise conservation property  \cite{bach-plos15, binder2016layer}. Consequently, neurons that contribute more to the ones in the following layers have higher scores. Authors in \cite{lapuschkin2016analyzing} extended this method to Fisher Vectors, showing how distinct models may consider certain regions of images as relevant or not. In addition, their conclusions shed doubts on the reliability of model decisions which can be biased by context.

\subsection{Reliability of Visualizations}
Deep neural networks are complex systems which are not fully understood yet  \cite{nguyen2015deep, goodfellow6572explaining, szegedy2013intriguing}. Authors in \cite{szegedy2013intriguing} and \cite{goodfellow6572explaining} investigated the counter-intuitive property called Adversarial Examples, which is further explored in this work. They showed that deep models decisions can be fragile, being susceptible to directed perturbations that are imperceptible to humans. The nonlinear nature of neural networks is presented as the main cause of such vulnerability.

Authors in \cite{nguyen2015deep} further analysed this subject, questioning the differences between patterns considered by humans and models. They create noisy images which received extremely high confidence model decisions. In this work, we build on their line of questions and ask if state-of-the-art explanation methods capture the actual importance of input features to a specific model, or if they trade this off for patterns that are more intuitive to humans. Authors in \cite{adebayo2018sanity} addressed this question and presented two tests to evaluate the reliability of explanation methods:

\begin{itemize}
\item \textbf{Model Parameter Randomization Test:} Compares the explanation maps of a trained model with the ones of a randomly initialized untrained network of the same architecture. If the results of the two cases do not differ significantly, it means that the explanation maps are insensitive to the parameters of the model.
\item \textbf{Data Randomization Test:} Compares the explanation maps of a model trained on a labeled dataset with the ones of a model with the same architecture but trained on a copy of the dataset in which the labels were randomly permuted. If the results do not differ significantly, the explanation maps are insensitive to the relationship between the input images and the original labels.
\end{itemize}

They evaluated widely used explanation methods, and concluded that visual inspection is a poor way of evaluating explanation results. These results imply the need for techniques which compare methods in a more objective manner, based on proper estimates of feature importance.

\subsection{Evaluation of Explanation Methods}

As stated above, comparisons between explanation methods of deep networks are commonly qualitative. They usually compare examples of relevance maps, preferring the ones more correlated to human expectation. In this sense, we consider the work in  \cite{samek2017evaluating} as the closest to ours. The authors provided a quantitative measure that evaluates how the relevant areas affect the correct prediction of a given model.
While they consider block regions of relevance and apply random changes to it, we use small guided perturbations in the whole input image. We evaluate the magnitude necessary for these perturbations to make the model change its output, considering that better relevance maps should require smaller magnitudes.

Other relevant work is  \cite{montavon2018methods}, in which the authors discussed the desirable properties of explanations and possible evaluation metrics, which they defined as follows:
\begin{itemize}
    \item \textbf{Explanation Continuity:} Ensures that if two inputs are nearly equivalent, then the explanations of their predictions should also be nearly equivalent.
    \item \textbf{Explanation Selectivity:} More relevant features should have stronger impacts on the classification. Thus, if features are attributed relevance, removing them should reduce evidence of the output.
\end{itemize}

Finally, authors in \cite{bach-plos15, samek2017evaluating} created concepts based on the quantification of this second property, measuring how fast an evaluated function starts to decrease when removing features with the highest relevance scores.

\section{APEM: Comparing Visual Explanations with Adversarial Perturbations}
\label{sec:apem}

We tackle the quantitative evaluation of explanations by performing guided perturbations to the original images. Typical explanation methods generate one relevance value for each pixel, resulting in images with the same dimensions of the input image, but in which values represent feature relevance $R = [[r_{i,j} | 0 \leq r_{i,j} \leq 1]]$. Thus, higher $r_{i,j}$ values imply those pixels have higher influence in the model decision.

\subsection{Assessing Reliability by Comparing Relevances} \label{sec:apem2}

Once a model is trained, the relevance maps can be calculated for a set of input images based on the model  parameters. Our aim is to produce scores that can be ranked to compare possible explanations for a given model, giving higher values to features that impact more the model decision.\footnote{APEM, however, does not produce results in the same scale for different models, so it cannot be straightforwardly used to compare explanations for different models with diverse capacities. Our objective is, thus, to compare different explanation methods given a fixed model.
}

We assume that changes in pixels associated with higher relevance values should impact the output more. In contrast, changing the irrelevant pixels should result in smaller impacts on the model outcome \cite{samek2017evaluating}. 
These perturbations are hardly noticeable and follow the directions of larger impact to the model decision, i.e. the models' gradients  \cite{goodfellow6572explaining}. We first normalize relevance values in $R$ by its $l_1$-norm:
$R_{norm} = \displaystyle\nicefrac{R}{norm(R)}$
\noindent so that explanation methods with different relevance scales could be directly compared. Then, we create a directed relevance $R_{dir}$ as show in Equation \ref{eq:dir_rel}. Specifically, for a model with parameters $\theta$, an input image $x$ and model output $\hat{y}=f_{\theta}(x)$ associated with $x$, we use the sign of the gradients from the loss function $J(\theta,x,\hat{y})$ to direct the relevances:
\begin{equation}\label{eq:dir_rel}
R_{dir} = R_{norm} \odot sign(\nabla_xJ(\theta,x,\hat{y}))
\end{equation}

\begin{figure}[t]
\centering
\includegraphics[width=\linewidth]{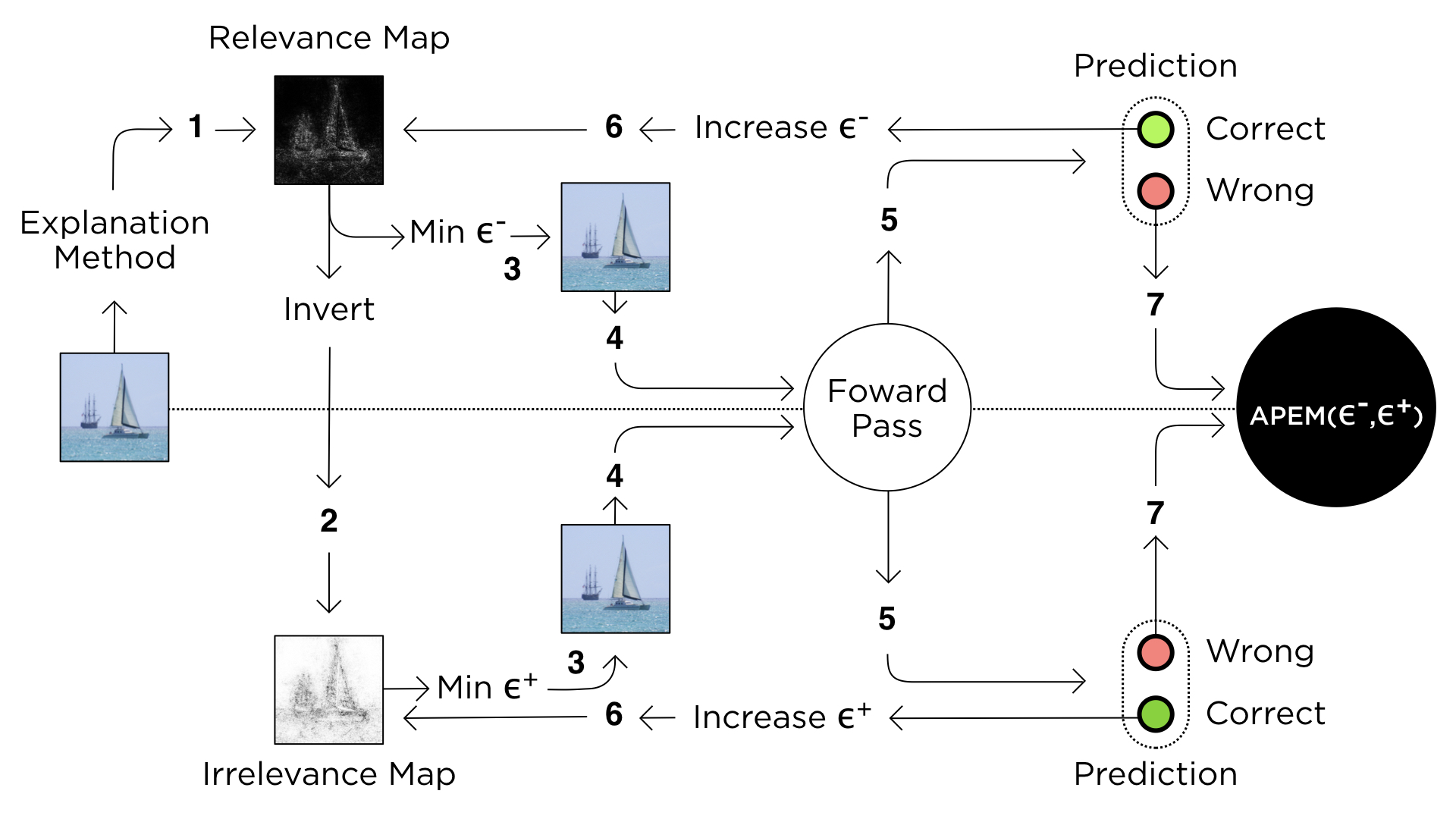}
\caption{Calculating APEM for an arbitrary input image. The transitions are enumerated sequentially from the first step to the last, and the $\epsilon$ values are stored and modified throughout the process. The diagram shows the relevance and irrelevance maps for an input image, and the gradient was used to create  relevance and irrelevance images. The final steps consist of perturbing the original image by using the maps until the model changes its output.}
\label{fig:diagram}
\end{figure}

Further, the $R_{dir}$ which correctly represents the model relevance would need a minimum perturbation to maximize the deviation from the model decision, as it is equivalent to one gradient ascent step in the pixel space. Therefore, there is a minimum $\epsilon^-$ value which makes the model change its prediction for the perturbed image $x'$ created from $x$:
$$
x' = x + R_{dir} \times \epsilon^- 
$$
Similarly, we argue that the values of irrelevance can analogously be extracted from $R$ by making $R^I = {1 - R}$. Moreover, the best ${R^I_{dir}}$ calculated from $R^I$ should take longer to change model decisions, as perturbations in irrelevant pixels should not cause significant changes in the output. This results in an $\epsilon^+$ value and, consequently, a gap between it and $\epsilon^-$. APEM is finally given as the average of all the gaps for a set of $n$ images, as expressed in Equation~\ref{eq:apem}. The entire process of computing APEM is depicted in Figure \ref{fig:diagram}.
\begin{equation}
\mbox{APEM} = \frac{\displaystyle\sum_{i=1}^n{ (\epsilon_i^+ - \epsilon_i^-) }}{n}
\label{eq:apem}
\end{equation}

\subsection{An Algorithm to Make Relevance Interpretable}
\label{sec:filter}

Since the reliability of a relevance map can now be measured, we may use APEM to assess if the quality of explanations was reduced when an explanation method simplifies relevance maps to make them more interpretable. We argue that a good simplification is one which does not reduce the APEM value of the original map. As discussed, pixels which are less relevant to the model decision have smaller values in the corresponding relevance map. When their relevance values are zeroed, their irrelevances are consequently set to $1$ and APEM can be calculated to the new map. If the APEM value remains unchanged, the pixels are considered to have no influence in the prediction and can be safely excluded from the explanation. This makes explanations more understandable, while keeping the same reliability levels.

We propose here, thus, a simple algorithm to filter relevance maps, removing the noise in it so humans can more easily interpret them. Basically, it zeroes the least relevant non-zero values in a relevance map iteratively, until any change in the relevance map causes a reduction in the APEM value. This algorithm can be applied to any explanation map calculated for an image and a trained model. While other methods have to deal with the trade-off of losing information to make explanations and visualizations more interpretable, the one presented here works as a way to enhance interpretation while not reducing explainability. Therefore, we have reliable visualizations and explanations which are also easier to understand.

\section{Experiments}
\label{sec:experiments}

In this section we report results from the comparison of different explanation methods using APEM. We also show how more interpretable visualizations may affect the actual explanation and how we can effectively tackle this problem, creating meaningful visualizations while keeping the same APEM values.

\subsection{Model, Data and Explanation Methods} \label{sec:data_and_methods}

Our model is a VGG-16 \cite{simonyan2014very} trained on the ILSVRC2012 dataset \cite{russakovsky2015imagenet} using PyTorch \cite{paszke2017automatic}. We used two sets of $5{,}000$ random images each -- one of correctly classified and another of misclassified images. These were taken from the validation set and used to create the relevances for each explanation method to be evaluated. We compare six explanation methods:

\begin{itemize}
\item \textbf{``Pure'' gradients}: the gradients are simply interpreted as a relevance map.
\item \textbf{Smooth Grad} \cite{smilkov2017smoothgrad}: it smooths the gradient, by applying a Gaussian kernel, instead of the raw gradient. This results in a sensitivity map $M$. Then, it averages the sensitivity maps using random samples obtained from the neighborhood of an input image $x$, formulated as:
$$ \hat{M}(x) = \frac{1}{n}\sum^n_1{M(x + \mathcal{N}(0, \sigma^2))} $$
where $n$ is the number of samples used, and $\mathcal{N}(0, \sigma^2)$ the Gaussian perturbation with standard deviation $\sigma$. 
\item \textbf{Layer-wise Relevance Propagation (LRP)} \cite{binder2016layer}: it computes the relevance of the pixels of an input image by considering their impact on the output of the model. LRP uses a graph structure to redistribute the relevance value at the output of the network back to the pixels. The relevance is propagated until it reaches the input, generating the pixel scores.
\item \textbf{Guided Backpropagation} \cite{springenberg2014striving}: it corresponds to the gradient method in which negative gradient entries are set to zero while backpropagating through a \textit{ReLU} unit.
\item \textbf{Grad-CAM} \cite{selvaraju2017grad}: it computes the relevance map as the gradient of the class score with respect to the feature map of the last convolutional unit of the network.
\item \textbf{Guided Grad-CAM} \cite{selvaraju2017grad}: Grad-CAM combined with Guided Backpropagation through an element-wise product for pixel-level granularity.
\end{itemize}

\begin{figure}[t]
\centering
   \includegraphics[width=.115\textwidth]{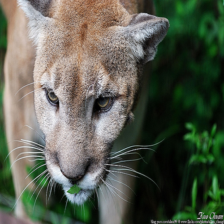}
   \includegraphics[width=.115\textwidth]{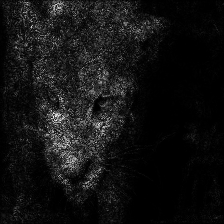}
   \includegraphics[width=.115\textwidth]{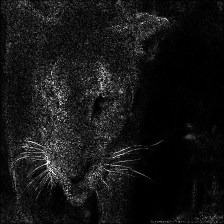}
   \includegraphics[width=.115\textwidth]{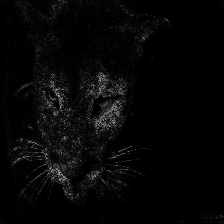}
   \includegraphics[width=.115\textwidth]{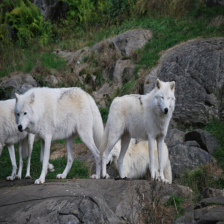}
   \includegraphics[width=.115\textwidth]{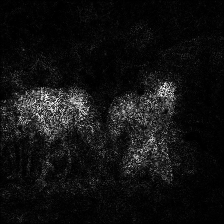}
   \includegraphics[width=.115\textwidth]{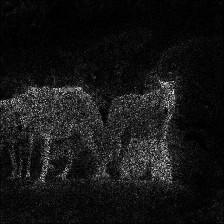}
   \includegraphics[width=.115\textwidth]{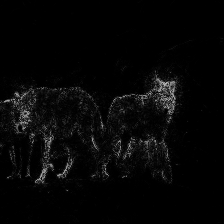}
   \includegraphics[width=.115\textwidth]{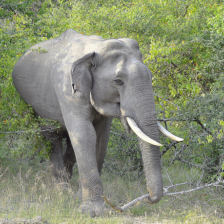}
   \includegraphics[width=.115\textwidth]{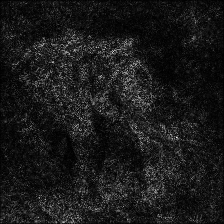}
   \includegraphics[width=.115\textwidth]{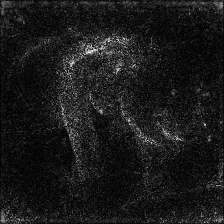}
   \includegraphics[width=.115\textwidth]{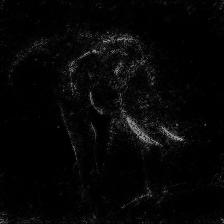}
  \caption{Examples of visualizations of relevance maps obtained by different explanation methods. Each row corresponds to an input image and each column shows a visualization: original image, and then the maps obtained with Gradient, SmoothGrad and LRP methods, respectively.}
  \label{fig:map_examples}
\end{figure}

For each of the aforementioned explanation methods, we get relevance values $R$ and clamp them to an upper-bound using its ninety-ninth percentile ($r_{i,j}'=\min(r_{i,j}, R_{99})$). Then, we multiply the new relevance values by their respective original image pixel values ($R''=R' \odot I$, where $\odot$ is the element-wise product).  This results in a cleaner visualization as proposed in \cite{smilkov2017smoothgrad}. As we are using RGB images, we reduce the number of channels in our relevance maps by summing them and normalizing it to the range $[0,1]$.\footnote{Though we present main results for all explanation methods described here, we will mostly focus on the Gradient, Smooth Grad and LRP methods throughout this work. We selected these three methods because of their different approaches and outcomes.}

\subsection{APEM Results}
\label{sec:results}

We start our analysis by showing a comparison between the different explanation methods using APEM values. Hyper-parameters used in Smooth Grad were $n = 100$ and $\sigma = 0.2$, and for LRP we used its $\epsilon\text{-variant}$ with $\epsilon = 1$.

\begin{figure}[t]
\centering
\includegraphics[width=.4\textwidth]{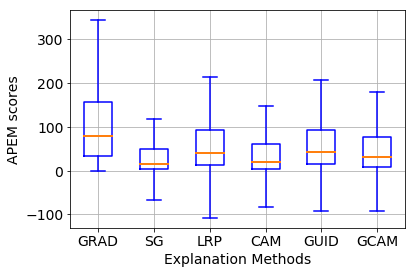}
\caption{Boxplots showing APEM values for each explanation method in the same set of images. They represent the Gradient, Smooth Grad, LRP, Grad-CAM, Guided Backpropagation and Guided Grad-CAM from left to right.}
\label{fig:average}
\end{figure}

Examples of the final maps are shown in Figure \ref{fig:map_examples}, for the Gradient, Smooth Grad and LRP methods. Although the relevance maps seem similar and easily interpretable, there are some particularities that are worth mentioning. In particular, while all visualizations present higher values in close regions, each visualization focuses in different parts of a same region. Figure \ref{fig:average} shows APEM  boxplots for each explanation method when considering all $5{,}000$ images. Higher APEM values mean better results, and clearly, the distribution of APEM values differs greatly depending on the explanation method. This indicates that while visualizations seem similar, they might not express the actual feature importance.

In order to assess the stability of the APEM values for each method across different network architectures, Table \ref{tab:res_results} presents the median of the results seen in the boxplots compared to a ResNet \cite{he2016deep} evaluated in the same conditions. We noticed the ranking of most explanation methods maintains the same order. The only exception is Grad-CAM and its guided variant, which show very similar results when evaluated on ResNet.

\begin{table}[t]
    \centering
    \caption{Median of the APEM values calculated for each explanation method for VGG and ResNet models. The relevance maps based on the Gradient result in the best scores. Implementing LRP on ResNet is not trivial, so we did not calculate the APEM score for this model.}
    \begin{tabular}{l | c c}
        \toprule
        & VGG & ResNet \\ %
        \midrule
        Gradient & \textbf{81.00} & \textbf{62.00} \\
        Smooth Grad & 16.00 & 15.00 \\
        LRP & 41.00 & -- \\
        Guided Bp & 43.00 & 28.00 \\
        Grad-CAM & 20.00 & 21.00 \\
        Guided Grad-CAM & 31.00 & 19.00 \\
        \bottomrule
    \end{tabular}
    \label{tab:res_results}
\end{table}

\begin{figure}[t]
\centering
\includegraphics[width=.4\textwidth]{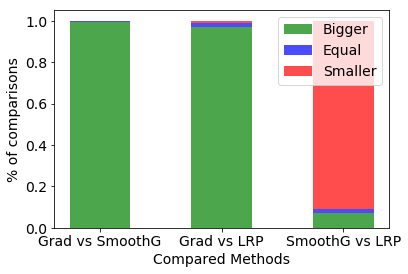}
\caption{Pairwise comparison between explanation methods. It shows the fraction of the total number of images in which one explanation method has a better/equal/worse APEM performance.}
\label{fig:pairwise}
\end{figure}

These results can be extended to a pairwise comparison, as shown in Figure \ref{fig:pairwise}. We counted the number of input images for which one explanation method beats the other in terms of APEM performance. Both Figures \ref{fig:average} and \ref{fig:pairwise} show LRP achieves a better APEM performance than SmoothGrad, but it also presents a higher standard deviation. This indicates that LRP is usually better than SmoothGrad, but it presents worse results for a few input images.

\subsection{Interpretable Visualizations vs. Actual Explanations}
\label{sec:intvsexp}

\begin{figure}[t]
\centering
    \includegraphics[width=.115\textwidth]{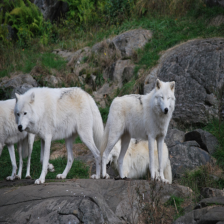}
    \includegraphics[width=.115\textwidth]{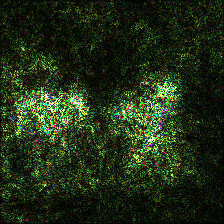}
    \includegraphics[width=.115\textwidth]{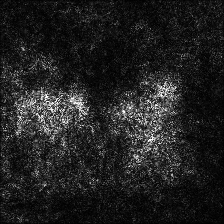}
    \includegraphics[width=.115\textwidth]{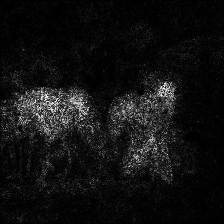}
    \includegraphics[width=.115\textwidth]{img/processing/ILSVRC2012_val_00000027_original.png}
    \includegraphics[width=.115\textwidth]{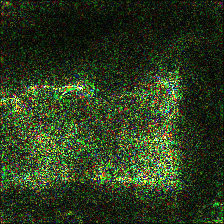}
    \includegraphics[width=.115\textwidth]{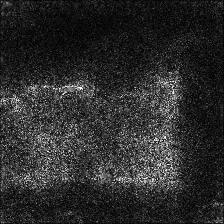}
    \includegraphics[width=.115\textwidth]{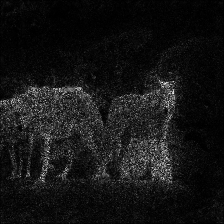}
    \includegraphics[width=.115\textwidth]{img/processing/ILSVRC2012_val_00000027_original.png}
    \includegraphics[width=.115\textwidth]{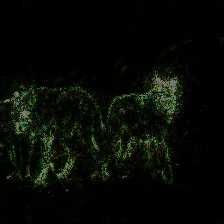}
    \includegraphics[width=.115\textwidth]{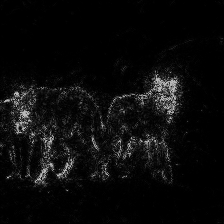}
    \includegraphics[width=.115\textwidth]{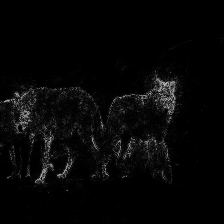}
\caption{Example of the relevance maps while filtering and simplifying visualizations. Lines correspond to the explanation method being used, which are the Gradient, SmoothGrad and LRP methods, from top to bottom. The filtering steps are presented in each column: (left) relevance in three channels, (middle) mapping into a single channel, and (right) relevance map multiplied by the original image.}
\label{fig:processing_vis}
\end{figure}

In this section we extend our discussion about the trade-off between a relevance map being interpretable and the actual importance that features within the relevance map have to model decisions. We address this problem by presenting the raw relevance values of a correct prediction and standard filtering steps that simplify the visualization until it becomes more comprehensible. First, relevance values are summed in the channel dimension so it becomes a relevance map. Then, they are multiplied by the grayscale image, so that it better fits the shapes in the original image, as proposed in \cite{smilkov2017smoothgrad}. Figure \ref{fig:processing_vis} shows examples of relevance maps obtained at each stage for the three explanation methods. Each method behaves differently -- some present great changes after each step, and for others only minimal changes are observed.

These filtering steps are likely to discard information that is important to the model. Indeed, APEM values decrease as filtering proceeds. Thus, we evaluate the loss of information that is lost during the filtering steps by using the average and median APEM values. Results are shown in Table   \ref{tab:processing_comp}, and they follow the same trend that was observed in Figure \ref{fig:processing_vis}: explanation methods that drastically change the relevance map during the filtering steps also present the greatest APEM losses, while methods that only produce minimal change are associated with the smaller APEM reductions.

\begin{table}[t]
    \centering
    \caption{APEM values calculated after each of the three simplification steps. APEM performance decreases as the simplification process proceeds.}
    \begin{tabular}{l | rrr rrr}
        \toprule
         & \multicolumn{3}{c}{Average} & \multicolumn{3}{c}{Median} \\
        \cmidrule(r){2-4} \cmidrule(r){5-7}
         & 1 & 2 & 3 & 1 & 2 & 3 \\ 
        \midrule
         Gradient & \textbf{231.76} & 176.86 & 127.79 & \textbf{137} & 105 & 81 \\
         Smooth Grad & \textbf{93.40} & 74.61 & 44.38 & \textbf{41} & 28 & 16 \\
         LRP & \textbf{101.65} & 93.93 & 66.07 & \textbf{59} & 51 & 41 \\         
         Guided Bp & \textbf{118.65} & 103.09 & 78.07 & \textbf{63} & 53 & 43 \\
         Grad-CAM & \textbf{100.86} & 100.86 & 42.25 & \textbf{41} & 41 & 20 \\
         Guided Grad-CAM & \textbf{83.00} & 69.95 & 37.79 & \textbf{51} & 42 & 31 \\
        \bottomrule
    \end{tabular}
    \label{tab:processing_comp}
\end{table}



Therefore, the application of these simplifications should be used considering the decrease in APEM performance compared to the amount of visual comprehension that they bring. Explanations which look more \emph{noisy} might be harder for an user to analyze than centered clouds of interpretable information, even if \emph{cleaning} the image means a loss in explanation. These factors should be weighted when applying explanation methods to practical applications. 

\begin{figure}[t]
\centering
    \includegraphics[width=.115\textwidth]{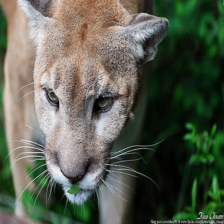}
    \includegraphics[width=.115\textwidth]{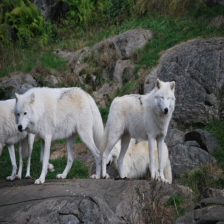}
    \includegraphics[width=.115\textwidth]{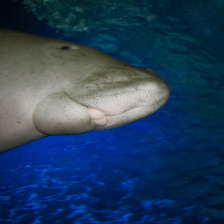}
    \includegraphics[width=.115\textwidth]{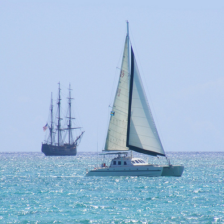}
    \includegraphics[width=.115\textwidth]{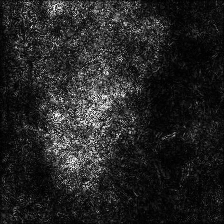}
    \includegraphics[width=.115\textwidth]{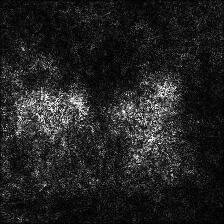}
    \includegraphics[width=.115\textwidth]{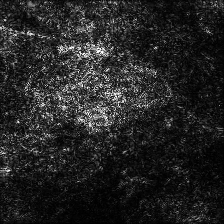}
    \includegraphics[width=.115\textwidth]{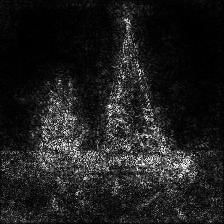}
    \includegraphics[width=.115\textwidth]{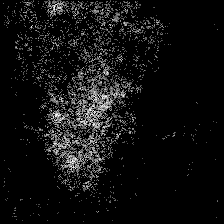}
    \includegraphics[width=.115\textwidth]{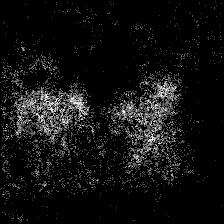}
    \includegraphics[width=.115\textwidth]{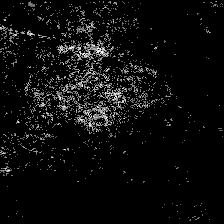}
    \includegraphics[width=.115\textwidth]{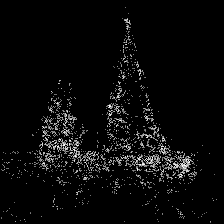}
\caption{Examples of the final images obtained with our filtering algorithm. The figure presents input images (top), their relevance maps based on the Gradient method before (middle) and after (bottom) the filtering process.}
\label{fig:filter_simple_map}
\end{figure}

\subsection{Filtering Explanations}

Next we expand the use of APEM to filter relevance maps so that they become more interpretable without any loss of essential information for the model, as we discussed in Section \ref{sec:filter}. Specifically, our objective is to make the visualization of relevance maps more interpretable as long as APEM values do not drop. In this case, the relevance maps are made more understandable, while still reliable -- in the sense that they comprise the features that actually impact model decision.

Figure \ref{fig:filter_simple_map} shows examples of relevance maps computed for the images and their last configurations before there is a drop in their APEM values. Interestingly, as the filtering steps proceed, regions outside the main object in the image were mostly erased from the relevance map. This means that the small relevance values attributed to the image's context were not actually relevant. For instance, the grass, stones and water present in the examples were given relevance values but they could be removed, leaving only the main objects in the images.

\begin{figure}[t]
\centering
    \includegraphics[width=.115\textwidth]{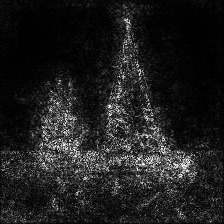}
    \includegraphics[width=.115\textwidth]{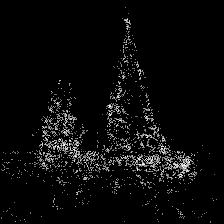}
    \includegraphics[width=.115\textwidth]{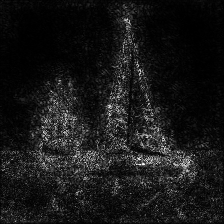}
    \includegraphics[width=.115\textwidth]{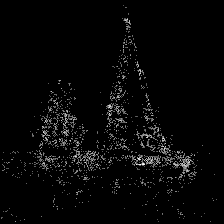}
    \includegraphics[width=.115\textwidth]{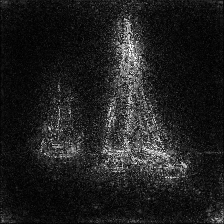}
    \includegraphics[width=.115\textwidth]{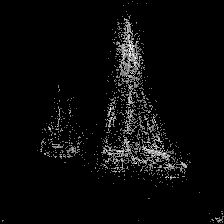}
    \includegraphics[width=.115\textwidth]{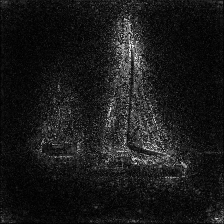}
    \includegraphics[width=.115\textwidth]{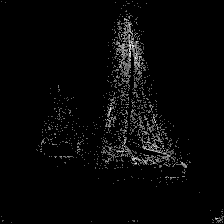}
    \includegraphics[width=.115\textwidth]{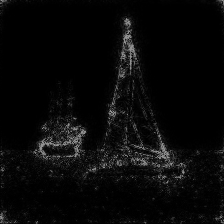}
    \includegraphics[width=.115\textwidth]{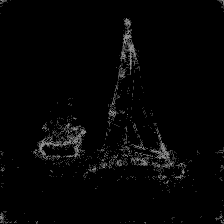}
    \includegraphics[width=.115\textwidth]{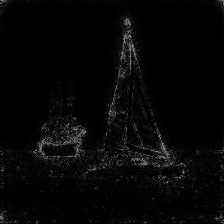}
    \includegraphics[width=.115\textwidth]{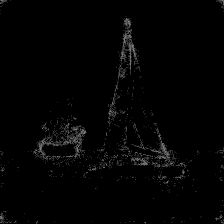}
\caption{Filtering algorithm applied to an image's relevance map. The rows correspond to the Gradient, SmoothGrad and LRP methods and the columns are, from left to right: (1) the relevance maps; (2) its filtered image; (3) the map multiplied by the original image; (4) its filtered image.}
\label{fig:filter_all}
\end{figure}

Our filtering algorithm can also be used within other explanation methods, as shown in Figure \ref{fig:filter_all}. Each map focuses on slightly distinct parts of the boat because of the characteristics of the explanation method, but all of them removed the relevance that was attributed to the sea, showing that this context was not considered relevant. Finally, the outcome of the filtering algorithm seems easier to interpret than the original noisy maps.

\section{Further Analysis}
\label{sec:analysis}

In this section we discuss interesting properties of APEM that could benefit future applications.

\subsection{The Importance of Irrelevance Maps}

\begin{figure}[t]
\centering
\includegraphics[width=.8\linewidth]{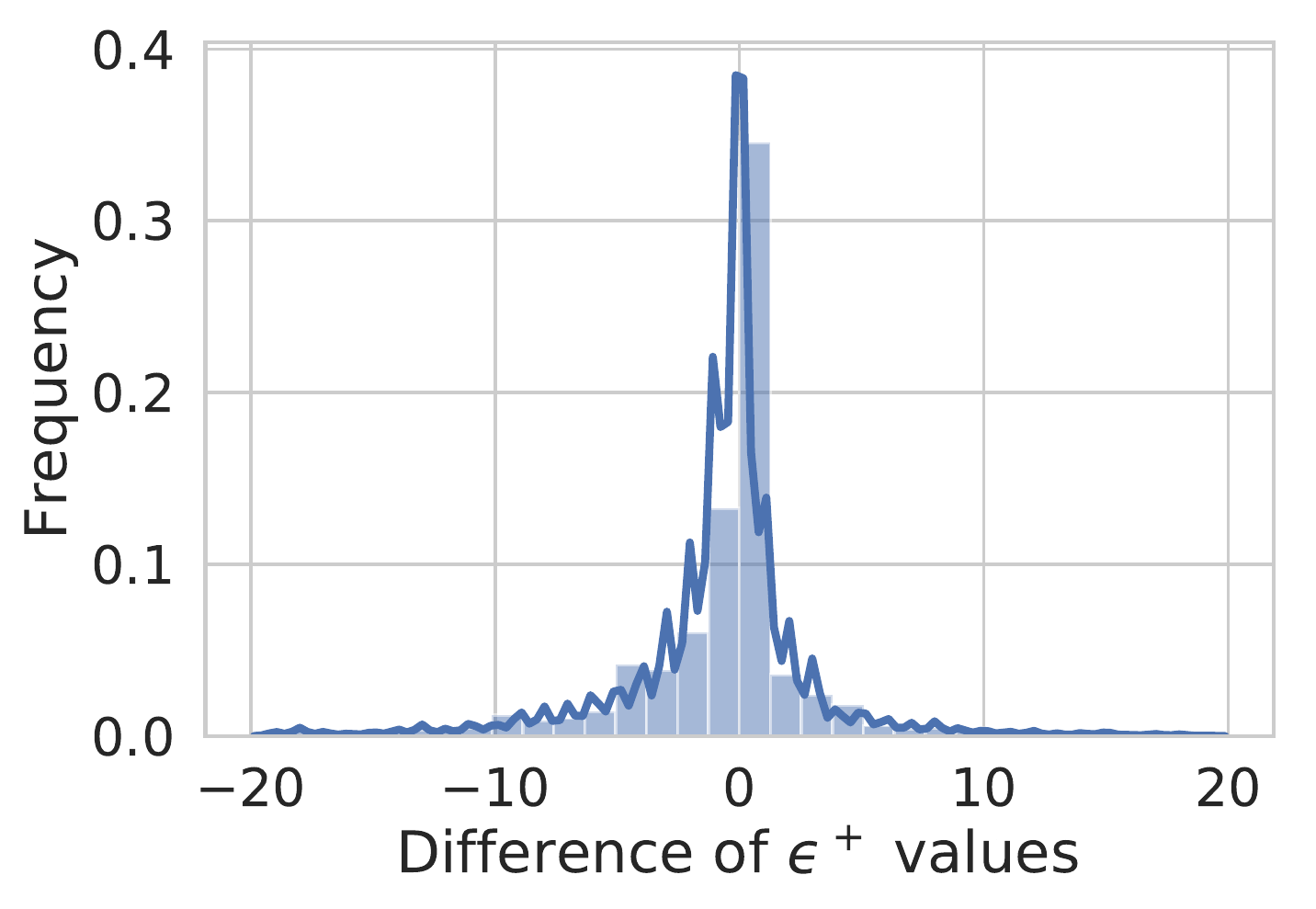}
\caption{Histogram and kernel density estimate of the difference between $\epsilon^+$ of LRP and SmoothGrad methods. Positive values are observed when $\epsilon^+$ for LRP is higher than for SmoothGrad.}
\label{fig:gap_vs_rel}
\end{figure}

As discussed previously, the inverse of relevance is also taken into account while calculating APEM. This is important because it prevents the explanation methods from focusing on a few relevant pixels, while not giving importance to others that may be also relevant. In this sense, considering irrelevance gives the metric a \emph{recall}-like property. In order to investigate the importance of also considering $\epsilon^+$ (irrelevance) when calculating APEM, we compared LRP with  SmoothGrad while only considering irrelevance values. Figure \ref{fig:gap_vs_rel} shows the histogram and the kernel density estimate of the difference between $\epsilon^+$ values from LRP and SmoothGrad. Although LRP presents higher APEM values than SmoothGrad, we observed that LRP produces relevance maps that are often more focused than those produced by SmoothGrad. LRP is, then, penalized for this and has lower $\epsilon^+$ values than SmoothGrad on average -- a lower $\epsilon^+$ results in a reduction in its APEM value.



\subsection{Single Instance Comparison}

\begin{figure}[t]
\centering
   \includegraphics[width=.23\textwidth]{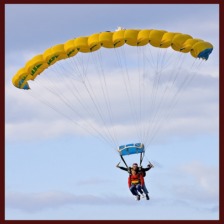}
   \includegraphics[width=.23\textwidth]{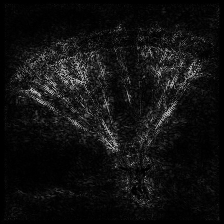}
   \includegraphics[width=.23\textwidth]{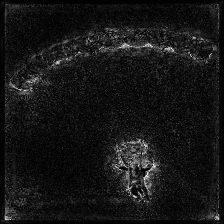}
   \includegraphics[width=.23\textwidth]{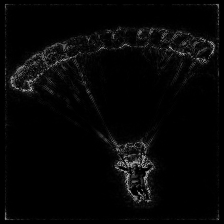}
  \caption{Example of an instance where the Gradient explanation results in worse APEM results than both LRP and SmoothGrad. Images represent: (top-left) original image, (top-right) Gradient, (bottom-left) SmoothGrad and (bottom-right) LRP.}
  \label{fig:bad_simple_grad}
\end{figure}

We analyze an (outlier) instance where the Gradient results in negative APEM values, which means its resulting irrelevance would be a better predictor of labeling choice than the relevance itself. For this specific image, shown in Figure \ref{fig:bad_simple_grad}, both LRP and SmoothGrad have better APEM values than the Gradient. In this instance, we can see that the Gradient focuses on the parachute strings and the clouds. Even though the gradient might be locally strong in that region, it is not a good predictor of class change, and, thus, not particularly relevant as an explanation of the model's labeling choice. At the same time, both LRP and SmoothGrad focus on the person and the parachute themselves, resulting in better APEM values.

%
%

\subsection{Misclassified Images}

So far our analysis only considered a set of $5{,}000$ correctly classified images. In order to properly explain and debug a model, we also have to understand how APEM behaves with misclassified images. Thus, we performed the same evaluation process on a set of $5{,}000$ misclassified images. In this case, the labels we used to create relevance maps were the ones predicted by the model.

As (hopefully) expected, the model is more uncertain about a decision when it predicts a wrong label. Then, it is expected that the model does not need many perturbations to make it change a prediction, which results in lower $\epsilon$ values. This leads to lower APEM values. On the other hand, this should not influence the quality of the explanation, disregarding the understanding of the generated relevance map. To put it simple, the overall APEM values are decreased but the best explanation methods should to keep their ranking positions.

Table \ref{tab:apem_results_wrong} shows average and median APEM values, comparing them with the ones obtained with the correctly classified images. The APEM decrease is clear, but the relative ordering of the explanation methods remains the same.

\begin{table}[t]
    \centering
    \caption{Average and Median APEM values for correctly classified and misclassified images. Relevance maps are calculated considering the prediction as the ground truth. Misclassification leads to an overall APEM decrease.}
    \begin{tabular}{l rr rr}
        \toprule
        & \multicolumn{2}{c}{Loss} & \multicolumn{2}{c}{Median} \\ \cmidrule(r){2-3} \cmidrule(r){4-5}
         & Correct & Misclassified & Correct & Misclassified \\ 
        \midrule
         Gradient & \textbf{127.79} & \textbf{43.58} & \textbf{81} & \textbf{25} \\
         Smooth Grad & 44.38 & 12.65 & 16 & 4 \\
         LRP & 66.07 & 20.34 & 41 & 11 \\ 
        \bottomrule
    \end{tabular}
    \label{tab:apem_results_wrong}
\end{table}

\subsection{Correlation between APEM and Loss}

The last set of experiments is devoted to investigate the possible correlation between APEM and loss. For this, we used the correctly classified images, misclassified images, and the total set of images. Again, the relevance map calculated for the misclassified images is based on the model prediction even though the loss uses the ground truth. The greatest probability for a label in the prediction is referred to as confidence, and it is also compared with the loss.

We compute the correlation using the Spearman's rank correlation coefficient \cite{croux2010influence} because of the non-linear relationships present in the data. This correlation is equal to the Pearson correlation between the rank values of the variables. A correlation close to $+$1 occurs when the observations have a similar rank between the variables, and it is close to $-$1 when they have a dissimilar one.

Table \ref{tab:corr} shows the correlations and the statistical significance. Our analysis indicates that correctly classified images have higher correlations while the misclassified images have virtually none. Also, observations have a dissimilar rank between APEM and loss, resulting in a negative correlation. Further, the methods that achieve higher APEM values also are the ones with higher correlation. In summary, the best explanation methods in terms of APEM have a higher correlation with the loss. Finally, high APEM values mean lower losses. Therefore, good explanations given by high APEM values may be used to assess the reliability of the model output.

\begin{table}[t]
    \centering
\caption{Spearman correlation between APEM values for each explanation method and the loss of the evaluated model ($\dag$ represents statistical significance with $\rho < 0.01$). The correlation of the confidence of the most probable label and the loss is also presented for comparison.}
    \begin{tabular}{l rrr}
        \toprule
        & \multicolumn{3}{c}{Loss} \\ \cmidrule(r){2-4}
         & Correct & Misclassified & Full \\ 
        \midrule
         Gradient & $\text{-}0.827^\dag$ & $0.069^\dag$ & $\text{-}0.543^\dag$ \\
         Smooth Grad & $\text{-}0.470^\dag$ & $\text{-}0.025$~~ & $\text{-}0.349^\dag$ \\
         LRP & $\text{-}0.602^\dag$ & $0.001$~~ & $\text{-}0.446^\dag$ \\
         Confidence & $\text{-}1.000^\dag$ & $\text{-}0.121^\dag$ & $\text{-}0.697^\dag$ \\
        \bottomrule
    \end{tabular}
    \label{tab:corr}
\end{table}

\section{Conclusions}
\label{sec:conclusion}

In this work, we proposed the Adversarial Perturbation Explanation Measure (APEM), a robust measure which evaluates the reliability of explanation methods. APEM enables us to compare explanation methods quantitatively, thus avoiding visual inspection. Moreover, it considers every relevance value for an input image to create perturbations and the irrelevance map to guarantee that no relevant pixel is left out. We present a comparison of some well-known explanation methods using our proposed measure. Along with it, we also present some characteristics of the methods and how APEM behaves. Furthermore, we showed some properties which especially make APEM robust. First, we showed the importance of using irrelevance as the result varies if the relevance map is completely precise but it is omitting other relevant pixels. Then, we analyzed the responses to misclassified images, showing that APEM drastically falls when the model is not able to correctly predict an instance. Finally, we correlated APEM results to the output of the model in different situations.

We also studied simplifications that aim to improve visualization of relevance maps. We showed the effect of these simplifications on the reliability of the resulting images, and a simple algorithm that works around this problem. The algorithm is one of the applications in which APEM can be used as a tool to filter the relevance maps into more interpretable images, while all the essential information is kept. The proposed algorithm can be used within an explanation method to create less noisy images and to facilitate its understanding.

\bibliographystyle{IEEEtran}
\bibliography{dan}

\end{document}